\documentclass[10pt,twocolumn,letterpaper]{article}

\usepackage[pagenumbers]{cvpr} %

\usepackage[accsupp]{axessibility}  %

\usepackage{graphicx}
\usepackage{amsmath}
\usepackage{amssymb}
\usepackage{booktabs}
\usepackage{multirow}
\usepackage[normalem]{ulem}
\useunder{\uline}{\ul}{}
\makeatletter
\newcommand{\thickhline}{%
    \noalign {\ifnum 0=`}\fi \hrule height 1pt
    \futurelet \reserved@a \@xhline
}
\makeatother
\usepackage{color}

\usepackage[pagebackref,breaklinks,colorlinks]{hyperref}

\usepackage[capitalize]{cleveref}
\crefname{section}{Sec.}{Secs.}
\Crefname{section}{Section}{Sections}
\Crefname{table}{Table}{Tables}
\crefname{table}{Tab.}{Tabs.}

\def\cvprPaperID{****} %
\def\confName{CVPR}
\def\confYear{2023}

\begin{document}

\title{Masked Images Are Counterfactual Samples for Robust Fine-tuning}

\author{Yao Xiao \qquad Ziyi Tang \qquad Pengxu Wei \thanks{Corresponding author.} \qquad Cong Liu \qquad Liang Lin \\
Sun Yat-sen University\\
{\tt\small \{xiaoy99,tangzy27\}@mail2.sysu.edu.cn \qquad \{weipx3, liucong3\}@mail.sysu.edu.cn} \\
{\tt\small linliang@ieee.org}
}
\maketitle

\begin{abstract}
Deep learning models are challenged by the distribution shift between the training data and test data. Recently, the large models pre-trained on diverse data have demonstrated unprecedented robustness to various distribution shifts. However, fine-tuning these models can lead to a trade-off between in-distribution (ID) performance and out-of-distribution (OOD) robustness. Existing methods for tackling this trade-off do not explicitly address the OOD robustness problem. In this paper, based on causal analysis of the aforementioned problems, we propose a novel fine-tuning method, which uses masked images as counterfactual samples that help improve the robustness of the fine-tuning model. Specifically, we mask either the semantics-related or semantics-unrelated patches of the images based on class activation map to break the spurious correlation, and refill the masked patches with patches from other images. The resulting counterfactual samples are used in feature-based distillation with the pre-trained model. Extensive experiments verify that regularizing the fine-tuning with the proposed masked images can achieve a better trade-off between ID and OOD performance, surpassing previous methods on the OOD performance. Our code is available at \url{https://github.com/Coxy7/robust-finetuning}.
\end{abstract}

\vspace{-10pt}
\section{Introduction}
\label{sec:intro}

\begin{figure}[t]
    \centering
    \includegraphics[width=0.45\textwidth]{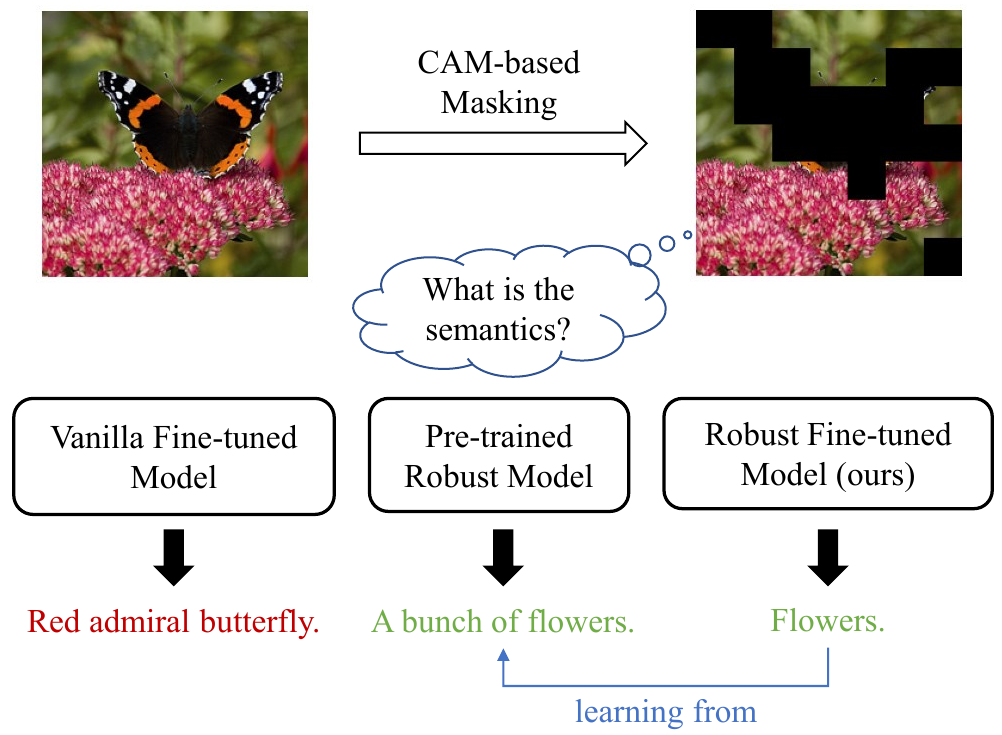}
    \caption{Illustration of our work. Vanilla fine-tuned models tend to learn spurious correlations that degrade the OOD robustness. To tackle this issue, our model learns from the pre-trained model on the counterfactual CAM-based masked images.}
    \vspace{-10pt}
    \label{fig:figure1}
\end{figure}

Deep learning has made impressive advances in various tasks on computer vision. Despite the remarkable performance achieved on benchmark datasets, deep models are often challenged by the distribution shift between the training data and test data  \cite{imagenet_v2,imagenet_r,imagenet_a,objectnet}. It is commonly assumed that the training and test samples follow the same distribution, which may not hold in real-world applications due to the unpredictable change of lighting conditions, viewpoints, backgrounds, \etc.
Although there are attempts to improve the robustness of deep models to the distribution shift (or OOD robustness), it is still rather under-explored \cite{miller2021accuracy,taori2020measuring}.

Recently, large-scale models pre-trained on diverse data have demonstrated unprecedented robustness to various distribution shifts \cite{clip,align,bommasani2021opportunities}.
Fine-tuning these pre-trained models on downstream tasks can be a promising approach to building robust models for different applications. However, it is found that while fine-tuning improves the performance on in-distribution (ID) data, it may reduce that on out-of-distribution (OOD) data \cite{lp_ft,wise}. %
To tackle this trade-off, several methods \cite{lp_ft,wise,model_soup} have been proposed to improve both ID and OOD performance in fine-tuning. However, they do not explicitly address the OOD robustness problem; instead, they implicitly preserve the robustness of the pre-trained model by constraining the distortion of pre-trained weights or using model ensembles. 

In this paper, we revisit the issue of robustness degradation in fine-tuning from a causal perspective. A large-scale pre-trained model somewhat shows properties in causality and stays robust to OOD samples \cite{willig2022can}.
However, when fine-tuning on downstream tasks, a majority of the model parameters tend to be adjusted for the downstream task in fine-tuning due to the highly entangled representation of images, arguably destructive to the generalizable knowledge \cite{scherrer2022generalization, ke2021systematic}.
In contrast, distribution shifts are usually sparse in the underlying causal factorization of the data generation process~\cite{bengio2019meta, scherrer2022generalization}. 
In this low-dimensional case, if we know which variables vary with different data distributions in this factorization (\ie, the non-stationary factors), we can achieve the OOD robustness by simply excluding their influence on the final predictions of the model. %

Specifically, we consider a Structural Causal Model (SCM) \cite{pearl2009causality} for the object-centric image generation process, as depicted in \cref{fig:causalgraph}. In this SCM, images are generated according to a non-stationary domain-relevant factor and a stationary semantic factor. Between them is a spurious correlation caused by a hidden non-stationary confounder that influences how the domain-relevant factor changes with the semantic one. To retain the OOD robustness, a fine-tuning model should avoid mapping non-stationary domain-relevant features to the predicted semantics.

To this end, we propose to fine-tune the models with masked images, which serve as counterfactual samples breaking the spurious correlation. Training on these samples helps preserve the stationary and generalizable knowledge of the pre-trained model. Concretely, we either mask the patches that contribute most to the label (\ie, the main object) or mask those with the least contribution (\eg, the context), which can be implemented based on class activation map (CAM) \cite{cam,chefer2021generic}. Such image masking forms an manipulation of a factual image and produces a counterfactual sample. Since the pre-trained model can better disentangle invariant features across domains, we require the fine-tuning model to learn from the pre-trained model on these counterfactual samples, as illustrated in \cref{fig:figure1}.
Furthermore, we argue that simply dropping the masked patches may be insufficient to alleviate the risk of fitting spurious correlations, and we propose to refill the masked patches with those from other images.

We study different combinations of masking strategies (\eg, masking the object or the context) and refilling strategies (\eg, filling with patches from single or multiple images). Experimental results suggest that most of the strategies are applicable to the construction of counterfactual samples that help improve the robustness in fine-tuning, while masking the object generally achieves the best robustness.
Compared with existing methods \cite{lp_ft,wise,model_soup} on fine-tuning CLIP \cite{clip} models, our approach achieves better average accuracy on various OOD datasets without relying on model ensembles or weight constraints.
We also find that, taking the weight-space ensemble of the zero-shot model and our fine-tuned model following WiSE-FT \cite{wise}, hardly improves the trade-off between ID and OOD accuracy, which contradicts previous observations and implies that our approach may produce essentially different models in comparison with conventional fine-tuning.

\section{Related Works}\label{sec:related_works}

\subsection{Causal Perspective on OOD Robustness}

Causality holds the promise of OOD robustness as it has the property of robustness under interventions that underlie distribution shifts~\cite{arjovsky2020out, scherrer2022generalization}. To learn the latent causal mechanism in visual data and attain OOD robustness, a branch of methods~\cite{lu2021invariant, khemakhem2020variational, Scholkopf2021TowardCR} target at learning domain-invariant causal representations and seek guarantees on generalization. Some approaches approximates this goal based on invariant risk minimization~\cite{arjovsky2019invariant, ahuja2020invariant}, risk extrapolation~\cite{krueger2021out}, adaptation speed~\cite{bengio2019meta}, or variational Bayes~\cite{liu2021learning}. Ilse \textit{et al.}~\cite{ilse2021selecting} help models escape from spurious correlations by training on handcrafted analog interventional data. In this work, we aim to explorethe preservation of the OOD robustness in pre-trained models equipped with generalizable knowledge during transfer, based on the analysis of an underlying data generation causal mechanism. 

\subsection{ID-OOD Trade-off in Fine-tuning}

While a series of large pre-trained models exhibit strong OOD robustness \cite{clip,align,bommasani2021opportunities}, empirical evidence suggests that fine-tuning these models on downstream data may degrade the robustness \cite{clip,lp_ft,wise}. It is also theoretically justified that compared with linear probing, vanilla fine-tuning may increase the ID performance while decreasing the OOD performance due to feature distortion \cite{lp_ft}.
To tackle this ID-OOD trade-off, LP-FT \cite{lp_ft} applies linear probing before fine-tuning; Calibrated Ensemble \cite{kumar2021calibrated} and  WiSE-FT \cite{wise} propose to take the ensemble of the pre-trained (zero-shot) model and vanilla fine-tuned model in output-space and weight-space, respectively. Model Soup \cite{model_soup} also improves the trade-off via weight-space ensembles, but it utilizes multiple models fine-tuned with different hyper-parameters.
While these methods are empirically effective in improving both ID and OOD performance, they only implicitly preserve the robustness of the pre-trained model by constraining the deviation of the downstream model from the pre-trained one.
Instead, our approach is based on explicit causal modeling of the OOD robustness problem and the use of counterfactual samples in fine-tuning.

\subsection{Learning with Masked Images}

Masked image modeling \cite{bao2021beit,mae,xie2022simmim,maskfeat} has been proven to be an effective approach to vision model pre-training, where random sampling is a common strategy for masking. In our task, we find that learning with CAM-based masking can better address the spurious correlation in images. 
Similar to our approach, CSS \cite{chen2020counterfactual} synthesizes counterfactual images for visual question answering (VQA) by masking out critical objects, for which the corresponding answer is changed to its negative; SwapMix \cite{gupta2022swapmix} swaps the context objects in the feature space to reduce the reliance of VQA models on visual context.  In this paper, we apply masking on images and refill the masked regions with content from other images, which is inspired by CutMix \cite{yun2019cutmix}. Besides, instead of manually constructing the labels for masked images like CSS or CutMix, we take the feature representations of the pre-trained model on the masked images as the supervision for the fine-tuning model.

\section{Method}

\subsection{Revisit OOD Robustness of Fine-tuned Models from a Causal Perspective} 
When fine-tuning a pre-trained model on downstream tasks in real-world applications, it is usually reliable to generalize only to in-distribution (ID) data. There are also increasing demands for robustness to distribution shifts or generalization to out-of-distribution (OOD) data. 
However, for a pre-trained large model that has been trained on data from diverse distributions and obtained generalizable knowledge (\eg, CLIP \cite{clip}), fine-tuning can lead to dramatic degradation of robustness \cite{lp_ft}.

This problem can be interpreted from a causal perspective. Suppose that we correctly factorize the data generation process in which variables are causally connected. Then, the discrepancy between ID and OOD (or sub-population) data generation process is usually sparse in this causal factorization~\cite{scherrer2022generalization, Scholkopf2021TowardCR}, \textit{e.g.}, few variables' priors change. The joint distribution that we can observe thereupon shifts. 
However, without modeling the causal hierarchy in data, the pre-trained models naturally learn dense and entangled features to represent the concepts in the causal factorization. Although distribution shifts are sparse in causal factorization, these models have to densely adapt their parameters in transfer to the downstream data~\cite{ke2021systematic}. In this case, the loss of generalizable knowledge is inevitable, which explains the significant decrease of robustness to distribution shifts. 

The core challenge in modeling the underlying causal factorization of visual data is the dense changes between ID and OOD data, despite the sparsity in causal factorization. For better robustness, a possible solution is to identify which visual features remain unchanged under distribution shifts, also known as \textit{stationary features}~\cite{scherrer2022generalization}. Without OOD test data in hand to identify stationary features, we seek to model the most important stationary features as the semantic features in the downstream object-centric task, and reduce non-stationary features to domain-relevant features. Finally, we establish our causal model as in Fig.~\ref{fig:causalgraph}.

\subsubsection{SCM for Image Generation Across Domains}

\begin{figure}[t]
    \centering
    \includegraphics[width=0.16\textwidth]{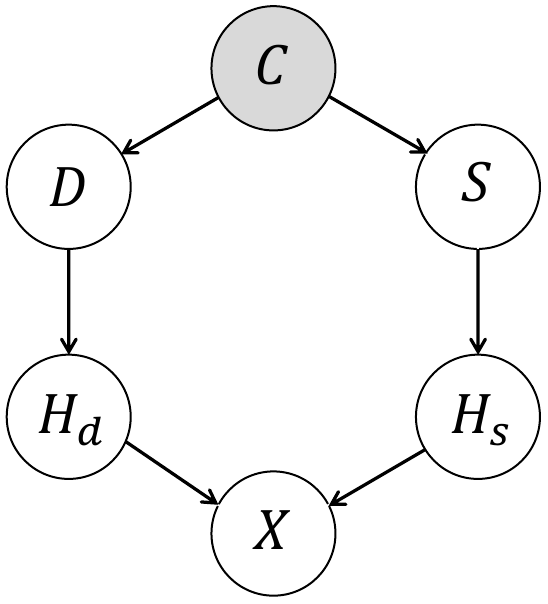}
    \vspace{-5pt}
    \caption{The causal graph of underlying object-centric image generation process across domains. $C$: confounder; $D$: domain; $S$: object semantics; $H_d$: (non-semantic) domain representation; $H_s$: semantic representation; $X$: image.}
    \label{fig:causalgraph}
    \vspace{-10pt}
\end{figure}

To illustrate the proposed method, we first introduce in Fig.~\ref{fig:causalgraph} a Structural Causal Model (SCM)~\cite{pearl2009causality} to model the underlying mechanism of object-centric image data generation across diverse domains, which is consistent with previous works~\cite{ilse2021selecting, subbaswamy2019preventing} despite some slight modifications. 

Formally, an SCM describes a directed acyclic graph composed of a set of endogenous variables $V$ and a set of exogenous variables $U$. An endogenous variable is a variable whose value is determined by other variables. Exogenous variables correspond to unobserved influences, usually considered independent of each other. In \cref{fig:causalgraph}, a set of endogenous variables are involved in the SCM, defined as $V = \{D, S, H_d, H_s, X\}$. There is also an exogenous variable for each endogenous variable, \textit{e.g.}, $U_{H_s}$ that possibly corresponds to the object pose and influences the generation of $H_s$. Here, we do not show them for simplicity, but only consider an important exogenous variable, the confounder $C$. We denote the set of exogenous variables by $U = \{C, U_D, U_S, U_{H_s}, U_{H_d}, U_X\}$. %

In this SCM instance, $C$ denotes an unobservable confounder variable, which is an exogenous variable shown in gray. For instance, it can be some specific time or space. $D$ is a domain variable containing information varied with the domain. $S$ is a semantic variable, \eg, the category of the object. In our setting, $D$ and $S$ are observable endogenous variables shown in white. Correspondingly, $H_d$ is the domain representation and $H_s$ is the semantic representation, \ie, they indicate how $D$ and $S$ are displayed in the image space, respectively. Finally, $X$ is the image generated by the interaction of $H_d$ and $H_s$.

\begin{figure*}[t]
    \vspace{-5pt}
    \centering
    \includegraphics[width=0.72\textwidth]{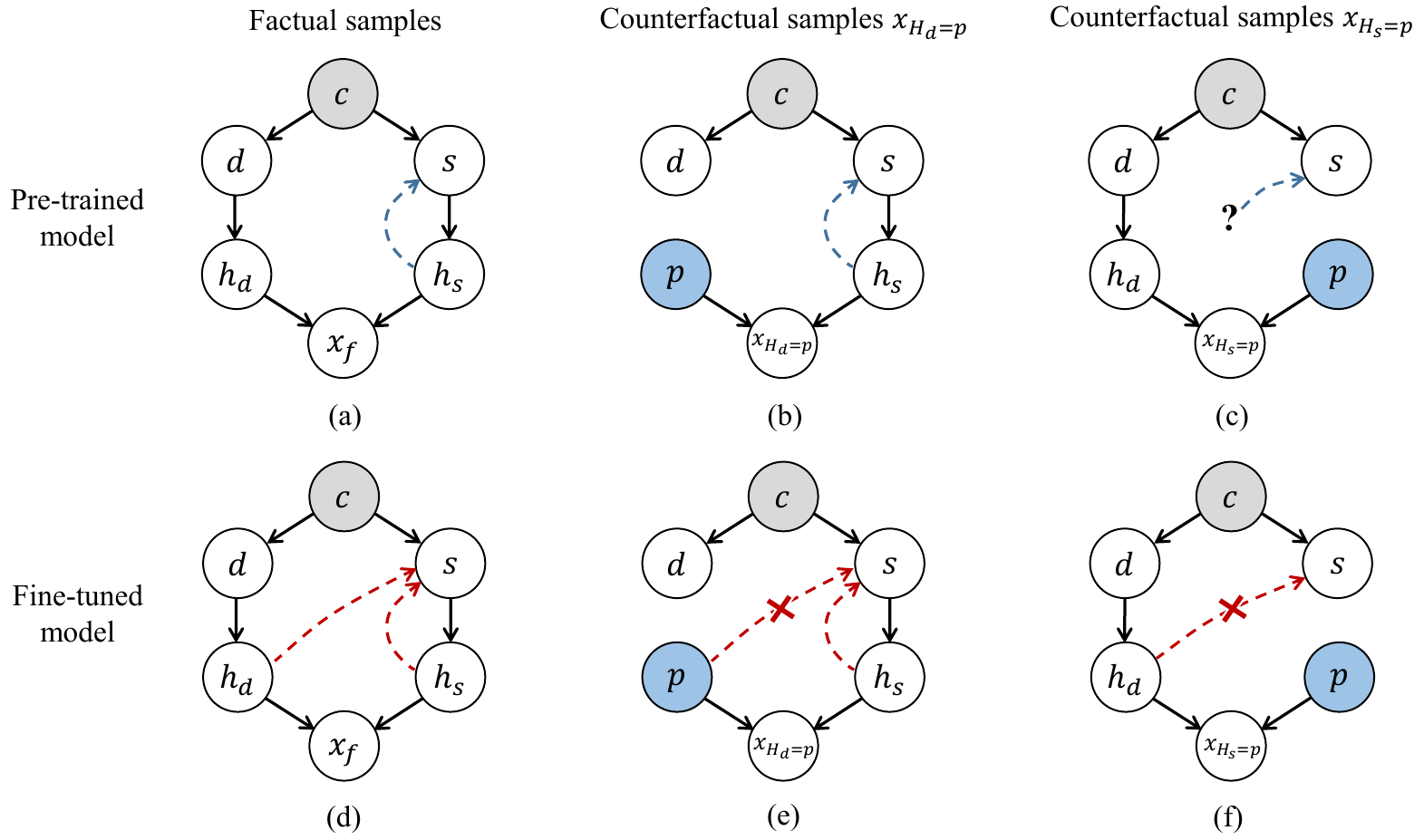}
    \vspace{-5pt}
    \caption{Conceptual comparison of pre-trained and fine-tuned models on what they depend on for predicting the semantic label of different samples. The robust pre-trained model is expected to depend only on $h_s$, while the fine-tuned model may associate both $h_d$ and $h_s$ to the semantics $s$.
    This may result in different predictions on counterfactual samples for the two models.
    }
    \label{fig:comparison}
\end{figure*}

\subsubsection{Spurious Correlation and OOD Robustness}

As depicted in \cref{fig:causalgraph}, there exists a backdoor path between $H_d$ and $H_s$, \textit{i.e.} $H_d \leftarrow D \leftarrow C \rightarrow S \rightarrow H_s$, making them spuriously correlated. Collecting data for a downstream task from a single source or environment may lead to strong spurious correlations between the semantic part $H_s$ and domain-relevant part $H_d$. In other words, there can be strong selection bias~\cite{lu2021invariant}. For example, in a downstream insect classification dataset, the \textit{red admiral butterflies} may all be captured sitting on flowers, as shown in \cref{fig:figure1}. However, this spurious correlation does not necessarily hold in OOD data~\cite{beery2018recognition}. Moreover, in reality, what makes the label \textit{red admiral butterfly} is the butterfly itself in the image space ($H_s$) rather than the flowers involved in $H_d$. Therefore, the model is required to depend only on $H_s$ for the prediction of $S$ to attain OOD robustness. 

For the reasons we have discussed earlier, even if a pre-trained large model has been endowed with reasonable knowledge to distinguish $H_s$ and $H_d$ (\cref{fig:comparison} (a)), the model is still susceptible to the spurious correlations during fine-tuning (\cref{fig:comparison} (d)). As a result, the transfer dramatically degenerates the pre-trained model's OOD robustness. 

From this perspective, a possible way to tackle the issue of robustness degradation is to break the spurious correlations in downstream data via specific manipulation, and explicitly requires the fine-tuning model to distinguish $H_s$ and $H_d$ following the pre-trained model. In the following sections, we propose a masking-based image manipulation method and indicate why it can be used to hinder the fine-tuning model from depending on $H_d$.

\subsection{Masked Images as Counterfactual Samples}
\label{sec:image_construction}

To break the spurious correlation in training images, we propose to mask out or replace certain regions of the images so that $H_d$ or $H_s$ is (partially) manipulated, which results in counterfactual samples.
Formally, given an observational sample $x$ and the SCM $M$, if we assign $d^\prime$ to variable $D$, the resultant counterfactual sample can be denoted as:
\begin{equation}
    \label{eq:counterfactual}
    x_{_{D=d^\prime}}(u) = x_{M_{D=d^\prime}}(u)
\end{equation}
where $M_{D=d^\prime}$ is the SCM instance in which $d^\prime$ is assigned to $D$, and $u$ denotes the values of the exogenous variables. %

We illustrate how the pre-trained and fine-tuned models may process two kinds of counterfactual samples differently in \cref{fig:comparison}.
First, for counterfactual samples whose domain-relevant representations $h_d$ are replaced by some other context $p$ (\ie, $x_{H_d=p}$ in \cref{fig:comparison} (b,e)), the fine-tuned model tends to attribute its prediction on both $p$ and $h_s$ since it has learned the spurious correlations. Hence, its prediction can be misled by $p$. Differently, the robust pre-trained model which can distinguish the semantic and non-semantic factors is not affected by $p$. 
Second, for counterfactual samples whose semantic representations $h_s$ are replaced by $p$ (\ie, $x_{H_d=p}$ in \cref{fig:comparison} (c,f)), the fine-tuned model can still predict the original semantics $s$ from $h_d$, while the pre-trained model cannot due to the missing semantic representations $h_s$.
In both cases, counterfactual samples may lead to different predictions between the two models.

Now the question is how to leverage these counterfactual samples to preserve the OOD robustness in fine-tuning. Directly training on these counterfactual samples with the labels of corresponding factual samples is unsuitable, since the original semantic information may be distorted.
Given that the pre-trained models can have substantial power to capture semantic cues, their image-level feature representations usually contain rich semantic information. We thereby seek an alternative solution in which these samples are used in distillation. This way, the fine-tuning model learns to mimic the pre-trained model in feature representation. Formally, we denote the image encoders of the pre-trained and fine-tuning model by $\hat{f}$ and $f$, respectively, and denote the classification head of the fine-tuning model by $g$. Then, the overall training objective can be written as follows:
\begin{equation}
    \label{eq:objective}
    \mathcal{L} = \mathcal{L}_\text{CE} \bigl ( g(f(x)) , y \bigr ) + \beta \mathcal{L}_\text{MSE} \bigl ( \hat{f}(x_{cf}) , f(x_{cf}) \bigr ),
\end{equation}
where $\mathcal{L}_\text{CE}$ is the cross-entropy loss, $\mathcal{L}_\text{MSE}$ is the mean squared error, $x$ is the raw image, $y$ is the label for $x$, $x_{cf}$ is the counterfactual image, and $\beta$ is a constant factor.

In the remaining parts of this section, we will discuss the proper design of masked images and show that it is nontrivial to construct the counterfactual samples $x_{cf}$ that can be used to effectively improve the OOD robustness.

\begin{figure}[t]
    \centering
    \includegraphics[width=0.48\textwidth]{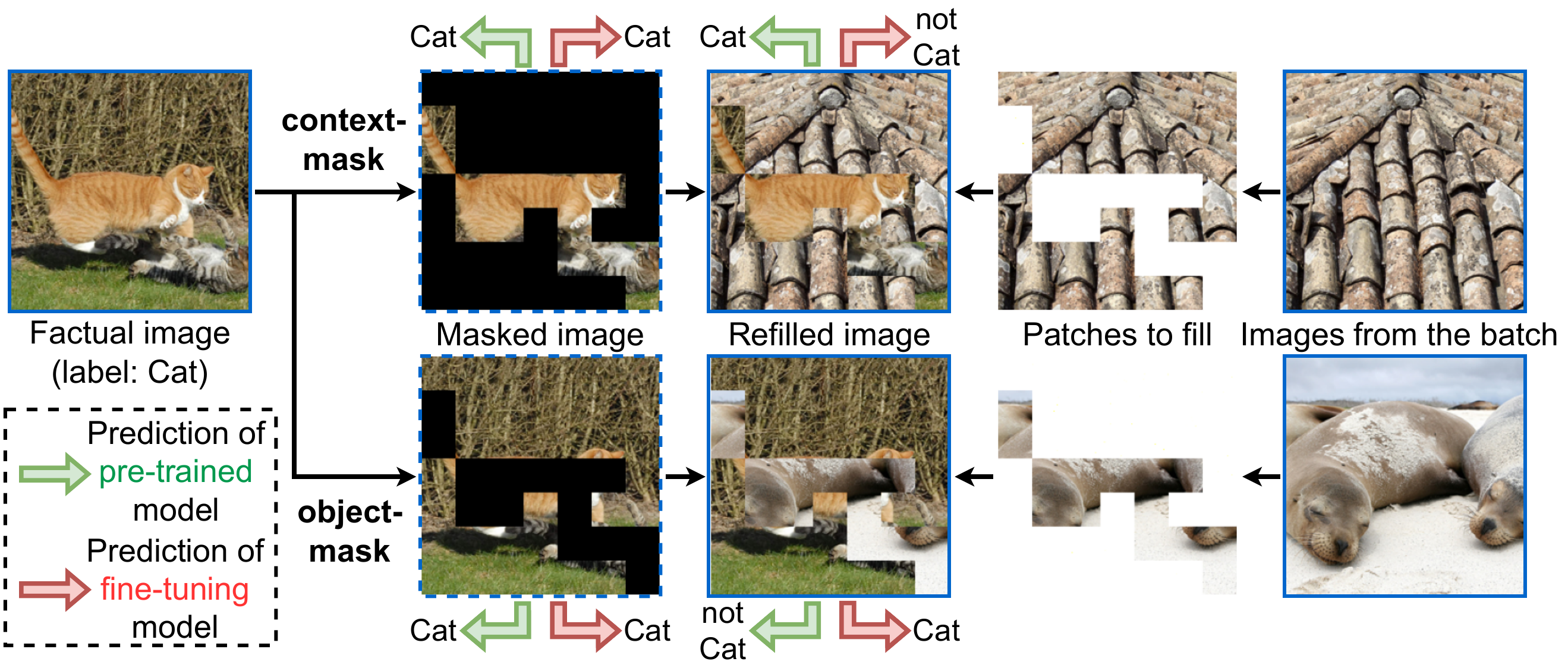}
    \vspace{-15pt}
    \caption{Illustration of the mechanism of masking and refilling.}
    \label{fig:refilling}
    \vspace{-10pt}
\end{figure}

\subsubsection{Choosing the Masked Regions} %
 
To generate counterfactual samples, we need to select the masked regions in an image where the contents can be manipulated.
Recent large-scale pre-trained models \cite{clip,align} usually adopt a Vision Transformer-based architecture \cite{vit}, dividing input images into a series of patches. Hence, we consider patch-based masking in this paper.
A basic strategy is to randomly sample a proportion of patches (\textit{random-mask}), which is adopted by recent pre-training methods like MAE \cite{mae}. However, random sampling is content-agnostic. Although both $H_s$ and $H_d$ may be manipulated in this scheme, it tends to leave some spurious correlations in the unmasked regions. It is found that an MAE model can achieve semantically plausible reconstruction from the randomly masked images with a high masking rate (\eg, 75\%), suggesting that such random masking may preserve a significant amount of the information of the whole images. Therefore, randomly masked images may be less suitable for improving OOD robustness. 

Alternatively, we investigate two content-based strategies that are complementary: (1) \textit{context-mask}: masking the patches that are least relevant to the label (\ie, $H_d$, usually the context); (2) \textit{object-mask}: masking the patches that are most relevant to the label (\ie, $H_s$, usually the main object). To measure the contribution of each image patch to the label, we generate the class activation map (CAM) \cite{cam,chefer2021generic} based on the fine-tuning model. Then, image patches are separated into two groups by a constant threshold $t$. In such a scheme, the semantic-relevant regions (where their activation values are larger than the threshold $t$) can coarsely serve as $H_s$, and non-semantic regions as $H_d$.

\subsubsection{Refilling the Masked Images}
\label{sec:refilling}

After choosing the patches to mask, simply dropping them following MAE\cite{mae} can lead to a counterfactual image, \ie, $x_{_{H_d=\emptyset}}(u)$. However, we argue that this strategy (abbreviated as \textit{no-fill}) can be insufficient for the construction of effective counterfactual samples for improving the robustness, and refilling the masked regions can tackle this issue.

Concretely, an effective counterfactual sample should cause contradictions between the pre-trained and fine-tuning model, so that the latter is regularized by the distillation loss in \cref{eq:objective} to depend less on non-semantic parts of images for the prediction of semantics.
However, masking without refilling may not construct such samples, as depicted in \cref{fig:refilling}.
Specifically, \textit{context-mask} may not produce contradictions since both models can predict semantics from the object, so we need refilling to bring some conflicting context that disturbs the prediction of the fine-tuning model.
\textit{Object-mask} alone can cause contradictions in theory, as the fine-tuning model could still predict the original semantics from the context, while the pre-trained model could not. However, due to the imperfect masking in practice, the pre-trained model may still recognize the object from its unmasked parts.
Then, refilling can further distort the original semantics to ensure contradictions.
Hence, refilling can help to construct more effective counterfactual samples for the proposed fine-tuning approach. %

\begin{figure}[t]
    \centering
    \includegraphics[width=0.39\textwidth]{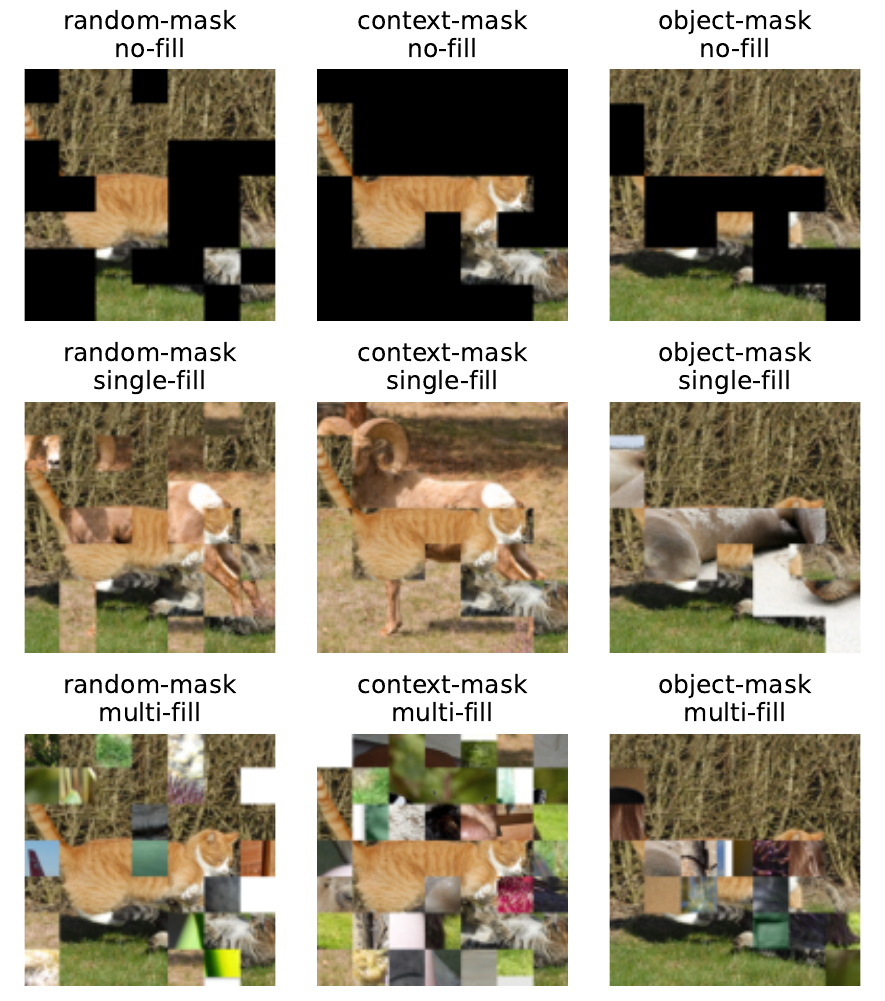}
    \vspace{-5pt}
    \caption{Illustration of different masking and refilling strategies. The raw (factual) image is displayed in \cref{fig:refilling}.}
    \vspace{-12pt}
    \label{fig:mask_mix_example}
\end{figure}

We consider two basic strategies to select the patches to fill: (1) \textit{single-fill}: select the patches in the corresponding positions of a single image randomly sampled from the training batch; (2) \textit{multi-fill}: for each patch to fill, independently sample a source image from the batch and select the patch in the corresponding position. All combinations of masking and refilling strategies are illustrated in \cref{fig:mask_mix_example}.

\section{Experiments}

\subsection{Setup}

\textbf{Datasets.}
We focus on fine-tuning the pre-trained model on ImageNet \cite{imagenet}, and evaluate the OOD robustness of the models on five datasets:
ImageNet-V2 \cite{imagenet_v2} is a new test set for ImageNet collected following the original protocol.
ImageNet-R \cite{imagenet_r} consists of various renditions (\eg, art, cartoons) of 200 ImageNet classes;
ImageNet-Sketch \cite{imagenet_s} contains sketches for each of the 1000 ImageNet classes;
ObjectNet \cite{objectnet} is a test set with 113 overlapping classes with ImageNet, which is collected to show objects from new viewpoints on new backgrounds;
ImageNet-A \cite{imagenet_a} contains natural images that are misclassified by models trained on ImageNet, covering 200 ImageNet classes.

\textbf{Evaluation.}
Following \cite{wise}, we use the top-1 accuracy as the metric of performance on ID and OOD data.
For the five OOD datasets, we report the top-1 accuracy on each dataset, as well as the average OOD accuracy computed by averaging the accuracy on the five datasets.
For datasets covering a subset of the ImageNet classes, the top-1 prediction of a model is taken as the class with the highest probability among the subset (instead of all ImageNet classes).

\textbf{Implementation details.}
Unless otherwise specified, we use the ViT-B/32 model \cite{vit} pre-trained via CLIP \cite{clip}.
The classification head of the zero-shot model is constructed from the pre-defined text prompts used by CLIP, and we adopt the implementation given by \cite{wise}.
To obtain accurate CAM scores for image masking, we apply the method proposed in \cite{chefer2021generic}, which is designed for attention-based models, including ViT.
For fine-tuning on ImageNet, we mainly follow the routine of WiSE-FT \cite{wise}. Specifically, we use the AdamW optimizer \cite{adamw} with a batch size of 512, and fine-tune for 10 epochs. The learning rate is set to $3\times 10^{-5}$ for all parameters and follows a cosine-annealing schedule \cite{loshchilov2016sgdr} with 500 warm-up steps. No data augmentation is applied apart from the necessary resizing and cropping, following the training of CLIP.
We split out a validation set of 10240 samples from the ImageNet training set to perform early stopping and model selection based on the validation accuracy.
$\beta$ in \cref{eq:objective} is fixed to be 30 for all experiments.
Additional details are in the appendix.

\subsection{Masking and Refilling Strategies}
\label{sec:exp_compare_strategies}

\begin{figure}[t]
    \centering
    \includegraphics[width=0.46\textwidth]{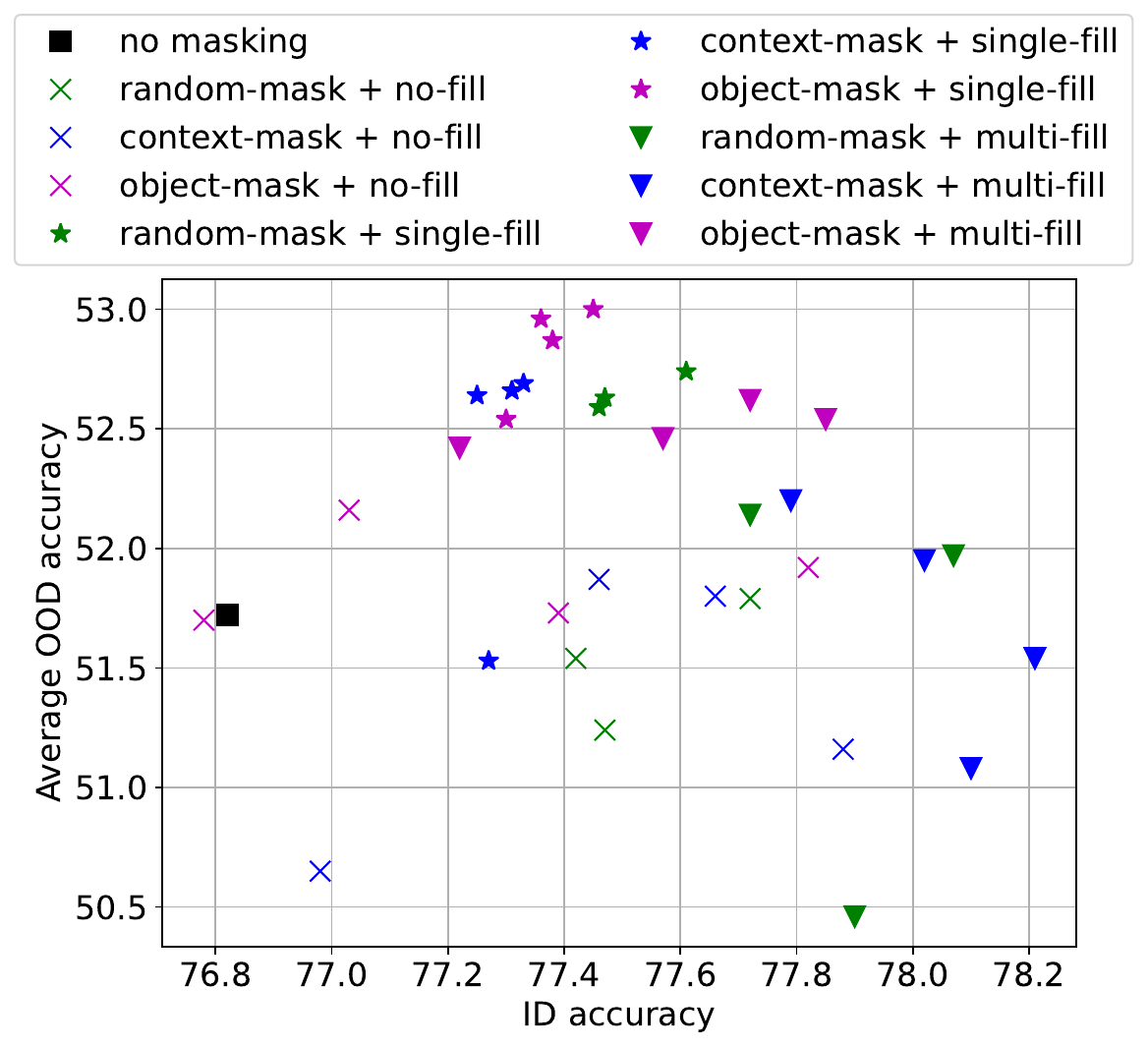}
    \vspace{-5pt}
    \caption{ID and average OOD accuracy of our approach using different masking and refilling strategies. For each combination of masking and refilling strategies, we plot multiple results based on different settings of hyper-parameters (\ie, masking rate or CAM score threshold). Best viewed in color.}
    \vspace{-12pt}
    \label{fig:compare_strategies}
\end{figure}

\begin{table*}[t]
\centering
\resizebox{\linewidth}{!}{
\setlength\tabcolsep{8pt}%
\begin{tabular}{c|cc|c|ccccc|c}
\thickhline
Method & Masking & Refilling & IN & IN-V2 & IN-R & IN-Sketch & ObjectNet & IN-A & OOD avg. \\ \hline
Zero-shot \cite{clip} & / & / & 63.4 & 55.9 & 69.3 & 42.3 & 44.5 & 31.4 & 48.7 \\
Vanilla fine-tuning & / & / & 75.9 & 64.7 & 57.0 & 39.8 & 39.5 & 20.0 & 44.2 \\
Ours & no masking & / & 76.9 & 66.5 & {\ul 69.2} & 45.6 & 45.3 & 29.8 & 51.3 \\ \hline
\multirow{3}{*}{Ours} & random-mask & \multirow{3}{*}{no-fill} & 77.5 & 66.9 & 66.4 & 45.7 & 46.5 & 30.8 & 51.2 \\
 & context-mask &  & 77.8 & 67.4 & 66.7 & 45.6 & 45.9 & 30.0 & 51.1 \\
 & object-mask &  & 77.7 & 67.1 & 67.6 & 46.2 & 46.8 & 31.5 & 51.9 \\ \hline
\multirow{3}{*}{Ours} & random-mask & \multirow{3}{*}{single-fill} & 77.6 & 67.1 & 69.0 & 46.4 & 47.8 & {\ul 33.4} & {\ul 52.7} \\
 & context-mask &  & 77.2 & 66.9 & 68.8 & 46.5 & {\ul 47.8} & 32.9 & 52.6 \\
 & object-mask &  & 77.5 & 67.1 & \textbf{69.7} & \textbf{46.9} & \textbf{48.0} & \textbf{33.8} & \textbf{53.1} \\ \hline
\multirow{3}{*}{Ours} & random-mask & \multirow{3}{*}{multi-fill} & {\ul 78.0} & 67.4 & 67.4 & 46.1 & 46.7 & 31.9 & 51.9 \\
 & context-mask &  & \textbf{78.2} & {\ul 67.4} & 66.5 & 45.5 & 45.9 & 30.0 & 51.1 \\
 & object-mask &  & 77.9 & \textbf{67.7} & 68.1 & {\ul 46.6} & 47.5 & 33.0 & 52.6 \\ \thickhline
\end{tabular}
}
\vspace{-5pt}
\caption{Comparison of different masking and refilling strategies. Accuracy on ImageNet (IN) and the five OOD datasets are reported. \textit{OOD avg.} is the average OOD accuracy on the five OOD datasets. The best accuracy is \textbf{bold-faced}, and the second best accuracy is {\ul underlined}. Results are averaged over three runs with different seeds.}
    \vspace{-5pt}
\label{tab:mask_mix}
\end{table*}

In this section, we validate the effectiveness of using masked images for improving OOD robustness, and investigate the different masking and refilling strategies discussed in \cref{sec:image_construction}.
Concretely, we consider three masking strategies:
(1) \textit{random-mask}: randomly select a fixed proportion of patches;
(2) \textit{context-mask}: masking the patches with CAM score lower than the threshold;
(3) \textit{object-mask}: masking the patches with CAM score higher than the threshold.
The masking rate for random-mask is selected from $\{0.25, 0.5, 0.75\}$, and the CAM score thresholds for the latter two strategies are selected from $\{0.3, 0.4, 0.5, 0.6\}$. 
The refilling strategies include:
(1) \textit{no-fill}: baseline strategy that drops the masked patches without refilling;
(2) \textit{single-fill}: refill with patches from one other image;
(3) \textit{multi-fill}: refill with patches from multiple other images.
We test all the combinations of masking and refilling strategies, and compare them with the baseline where no masking is applied (\ie, $x_{cf}=x$ in \cref{eq:objective}).

In \cref{fig:compare_strategies}, we plot the average OOD accuracy against ID accuracy for each model trained with the above strategies.
Besides, for each combination of masking and refilling strategies, we select the model with the highest validation accuracy, and report their results on each dataset in \cref{tab:mask_mix}.
Based on these results, we have the following conclusions.
(1) Most of the combinations of masking and refilling strategies achieve better ID-OOD trade-off than the no-masking baseline, which suggests the effectiveness of image masking in our proposed approach.
(2) Refilling masked images with patches from other images (\ie, single-fill or multi-fill) is better than solely dropping the masked patches.
(3) Comparing the two proposed refilling strategies, single-fill generally results in better OOD accuracy, while multi-fill may yield better ID accuracy.
(4) Comparing the masking strategies, object-mask is generally better than random-mask and context-mask in terms of OOD accuracy.

The superiority of object-mask can be explained following our analysis in \cref{sec:refilling}. Concretely, while both masking strategies are theoretically valid, context-mask is more dependent on refilling, since it supposes that the refilled context can effectively lead to contradictions. This may not be satisfied by the proposed refilling strategies, as they are not aware of the content of the patches taken from other images. Hence, context-mask can be less effective in practice.

\begin{table}[t]
\centering
\resizebox{\linewidth}{!}{
\setlength\tabcolsep{8pt}%
\begin{tabular}{c|cccccc}
\thickhline
Threshold & 0.7 & 0.6 & 0.5 & 0.4 & 0.3 & 0.2 \\ \hline
Image MR & 0.06 & 0.10 & 0.15 & 0.25 & 0.39 & 0.55 \\
Object MR & 0.17 & 0.23 & 0.33 & 0.48 & 0.65 & 0.80 \\
IoU & 0.15 & 0.21 & 0.28 & 0.38 & 0.45 & 0.48 \\
\thickhline
\end{tabular}
}
\vspace{-5pt}
\caption{Validation of CAM-based object masking with different thresholds. Image masking rate (MR): masking rate of the whole image. Object masking rate (MR): masking rate concerning the object area. IoU: Intersection over Union regarding the object. Averaged over a subset of masked samples constructed in training.}
\vspace{-10pt}
\label{tab:mask_rate}
\end{table}

In addition, to verify that the superiority of our CAM-based object masking is due to the effective location of objects, we calculate the average masking rates of the objects during training. Specifically, we evaluate the object-mask and single-fill strategy with different CAM score thresholds.
To obtain the masking rate of the object for an ImageNet image, we take the pixel-level mask produced by a pre-trained segmentation model \cite{cheng2022masked} as the approximation of ground truth.
In addition, we compute the masking rate of the whole image and the Intersection over Union (IoU) regarding the object.
Details are provided in the appendix.
As shown in \cref{tab:mask_rate}, the object masking rate is significantly higher than the masking rate of the whole image, which suggests that the object-mask mainly masks the objects. Besides, as the threshold for masking decreases, both the object masking rate and IoU increase.
Hence, the CAM-based object masking is empirically sound.

\subsection{Comparison with Existing Approaches}

\begin{table*}[]
\centering
\resizebox{\linewidth}{!}{
\setlength\tabcolsep{11pt}%
\begin{tabular}{c|c|c|ccccc|c}
\thickhline
Model & Method & IN & IN-V2 & IN-R & IN-Sketch & ObjectNet & IN-A & OOD avg. \\ \hline
\multirow{6}{*}{\begin{tabular}[c]{@{}c@{}}CLIP\\ ViT-B/32\end{tabular}} & Zero-shot \cite{clip} & 63.4 & 55.9 & 69.3 & 42.3 & 44.5 & 31.4 & 48.7 \\
 & Vanilla fine-tuning & 75.9 & 64.7 & 57.0 & 39.8 & 39.5 & 20.0 & 44.2 \\
 & WiSE-FT$^\dagger$ \cite{wise} & 76.6 & 66.6 & \textbf{70.2} & {\ul 47.1} & 46.3 & 31.9 & 52.4 \\
 & Uniform soup$^\ddagger$ \cite{model_soup} & \textbf{80.0} & \textbf{68.6} & 66.6 & \textbf{47.7} & 46.1 & 29.2 & 51.6 \\
 & Ours (multi-fill) & {\ul 77.9} & {\ul 67.7} & 68.1 & 46.6 & {\ul 47.5} & {\ul 33.0} & {\ul 52.6} \\
 & Ours (single-fill) & 77.5 & 67.1 & {\ul 69.7} & 46.9 & \textbf{48.0} & \textbf{33.8} & \textbf{53.1} \\ \hline
\multirow{6}{*}{\begin{tabular}[c]{@{}c@{}}CLIP\\ ViT-B/16\end{tabular}} & Zero-shot \cite{clip} & 68.3 & 61.9 & 77.6 & 48.3 & 54.0 & 50.1 & 58.4 \\
 & Vanilla fine-tuning & 80.7 & 70.4 & 64.0 & 45.1 & 49.1 & 35.2 & 52.8 \\
 & LP-FT \cite{lp_ft} & 81.7 & 71.6 & 72.9 & 48.4 & / & 49.1 & / \\
 & WiSE-FT \cite{wise} & 81.7 & 72.8 & \textbf{78.7} & \textbf{53.9} & {\ul 57.3} & {\ul 52.2} & {\ul 63.0} \\
 & Ours (multi-fill) & \textbf{82.5} & {\ul 73.4} & 76.4 & 52.7 & 56.8 & 52.0 & 62.3 \\
 & Ours (single-fill) & {\ul 82.4} & \textbf{73.4} & {\ul 78.1} & {\ul 53.4} & \textbf{57.9} & \textbf{53.5} & \textbf{63.3} \\ \thickhline
\end{tabular}
}
    \vspace{-5pt}
\caption{Accuracy of different methods for fine-tuning CLIP models on ImageNet (IN). \textit{OOD avg.} is the average OOD accuracy on the five OOD datasets. The best accuracy is \textbf{bold-faced}, and the second best accuracy is {\ul underlined}. Our results are averaged over three runs with different seeds. ($\dagger$: our implementation. $\ddagger$: our evaluation on the official model.)
}
    \vspace{-5pt}
\label{tab:sota}
\end{table*}

We compare our method with three existing approaches, namely LP-FT \cite{lp_ft}, WiSE-FT \cite{wise} and Model Soup \cite{model_soup}, which also aim to improve both ID and OOD accuracy of fine-tuning.
For WiSE-FT, we report the results with the default mixing coefficient, \ie, $\alpha=0.5$.
For Model Soup, we take the \textit{uniform soup} as the representative method, which achieves the best average OOD accuracy on CLIP fine-tuning as reported in \cite{model_soup}.
For our approach, we consider both refilling strategies, and adopt the object masking with the CAM score threshold that yields the highest validation accuracy.
The results are shown in \cref{tab:sota}.
\textbf{First}, our approach surpasses the previous approaches in terms of average OOD accuracy.
\textbf{Second}, our approach achieves better ID accuracy than other approaches except for Model Soup. Note that Model Soup takes the ensemble of many fine-tuned models with different hyper-parameters, including those trained with strong data augmentation. Conversely, we do not dive into hyper-parameters tuning and do not use data augmentation in our experiments.
\textbf{Third}, compared with WiSE-FT, our approach achieves superior performance on ObjectNet and ImageNet-A, but is inferior on ImageNet-R and ImageNet-Sketch, which suggests that the two approaches may improve different perspectives of robustness.
This motivates us to consider the integration of WiSE-FT. %

\subsection{Integrating WiSE-FT}

\begin{figure}[t]
    \centering
    \includegraphics[width=0.44\textwidth]{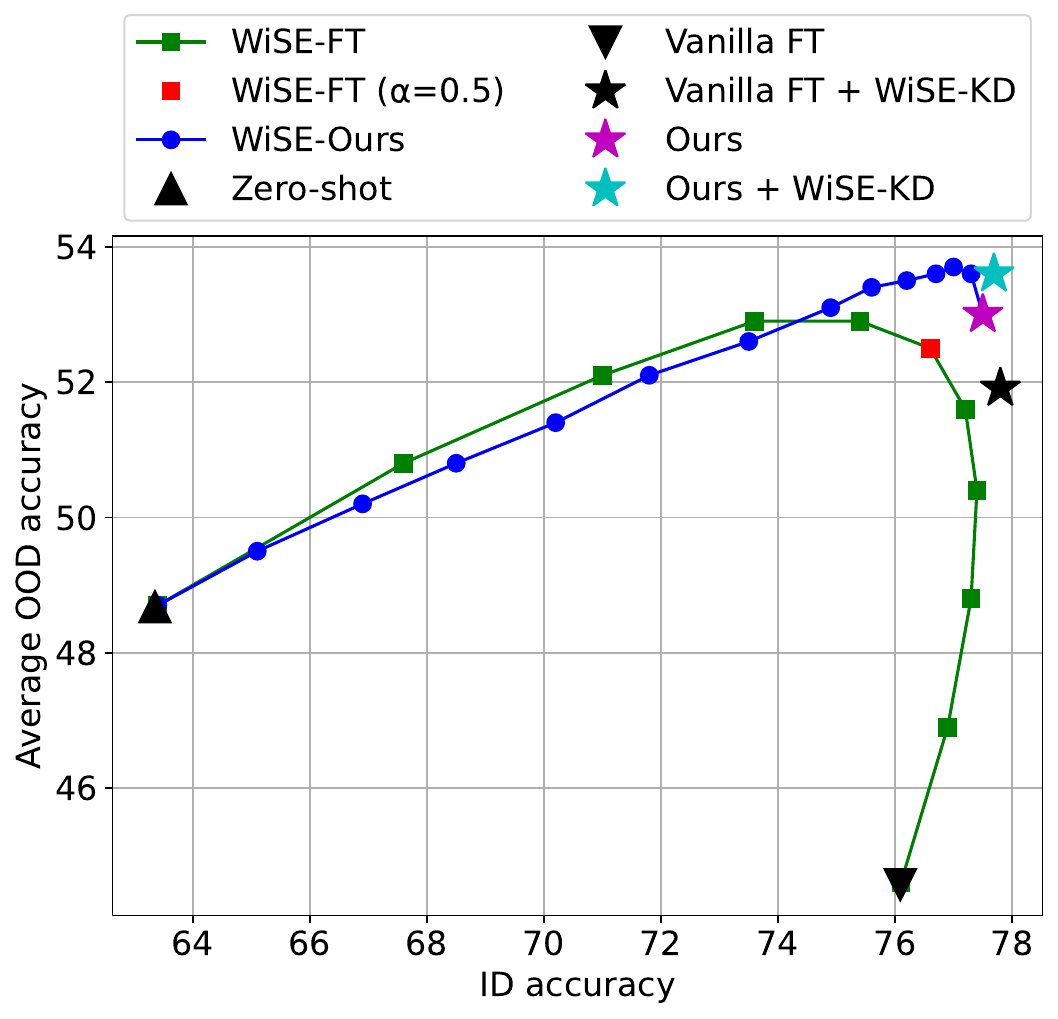}
    \vspace{-5pt}
    \caption{Integration of WiSE-FT and our approach. WiSE-Ours: weight-space ensembles of the zero-shot model and our fine-tuned model. WiSE-KD: knowledge distillation from the WiSE-FT model ($\alpha=0.5$). Vanilla FT: vanilla fine-tuning. For WiSE-FT and WiSE-Ours, each point corresponds to a value of $\alpha\in[0,1]$.
    }
    \vspace{-10pt}
    \label{fig:integrate_wise}
\end{figure}

WiSE-FT ensemble the weights of the zero-shot model and the vanilla fine-tuned model, and the ensemble model may achieve better accuracy on both ID and OOD data. Concretely, the weight ensemble is computed by
$\theta_e = (1-\alpha) \cdot \theta_0 + \alpha \theta_1$,
where $\theta_0$ is the weights of the zero-shot model, and $\theta_1$ is the weights of the fine-tuned model.
This weight-space ensemble method is agnostic of the fine-tuning method in form. Therefore, we may expect that the ensemble of the zero-shot model and the model fine-tuned with our method can achieve even better results.
However, as shown in \cref{fig:integrate_wise}, such direct integration with WiSE-FT may not improve ID and OOD accuracy simultaneously. Specifically, only for $\alpha$ close to 1, the resulting model may achieve better average OOD accuracy, but still at the cost of lower ID accuracy. This is different from the observations of WiSE-FT that using a medium value of $\alpha$ (\eg, 0.5) usually yields near-optimal results and surpasses the fine-tuned model on both ID and OOD accuracy \cite{wise}.
It is suggested that our proposed method may produce a significantly different model as compared with vanilla fine-tuning. In other words, our fine-tuned weights may leave the basin of the loss landscape where the weights of most vanilla fine-tuned models and the pre-trained model lie in \cite{neyshabur2020being,model_soup,wise}.

Another way to integrate WiSE-FT into our method is to utilize the ensemble model produced by WiSE-FT to guide the fine-tuning model via knowledge distillation (KD) \cite{kd}. For brevity, we name this method ``WiSE-KD''. Specifically, we consider using the WiSE-FT model as another teacher model, and add the vanilla knowledge distillation loss \cite{kd} to our training objective \cref{eq:objective}.
Details are presented in the appendix.
As presented in \cref{fig:integrate_wise}, applying WiSE-KD to our approach slightly improves both ID and OOD accuracy. As a comparison, while applying WiSE-KD to vanilla fine-tuning results in significantly better performance, it is still inferior to our approach, especially for OOD accuracy. This again validates the effectiveness of our approach in improving the OOD robustness in fine-tuning.

\section{Conclusion and Limitation}

In this paper, we analyze the issue of robustness degradation in fine-tuning from a causal perspective, and find that masked images can be effective counterfactual samples for breaking the spurious correlation between semantic and non-semantic factors and improving the OOD robustness of fine-tuning models. Experiments suggest that our approach surpasses previous methods on the OOD performance with competitive ID performance.

\textbf{Limitations.}
As stated in \cref{sec:exp_compare_strategies}, our refilling strategies are unaware of the content of the patches to fill, which limits the effectiveness of the resulting counterfactual samples. Besides, the feature-based distillation may not regularize the learning of the classification head of the fine-tuning model.
Future works can devote to better refilling and distillation methods conforming to the causal modeling.

\section*{Acknowledgments}
This work was supported in part by NSFC (No. U21A20470), National Key R\&D Program of China (No. 2021ZD0111601), and the Science and Technology Program of Guangzhou, China (No. 202201011550).

{\small
\bibliographystyle{ieee_fullname}
\bibliography{ref}
}

\appendix

\section{Experiment Details}

\subsection{Training Routines}

For fine-tuning on ImageNet via vanilla fine-tuning or our approach, we use the AdamW optimizer \cite{adamw} with $\beta_1=0.9$, $\beta_2=0.999$, weight decay of 0.1 and gradient clipping at $\ell_2$-norm 1. We use a batch size of 512, and fine-tune for 10 epochs. The learning rate is set to $3\times 10^{-5}$ for all parameters and follows a cosine-annealing schedule \cite{loshchilov2016sgdr} with 500 warm-up steps.
For both training and testing, we resize and center-crop the images to the size of $224\times 224$, and no data augmentation is applied.
Besides, different from WiSE-FT \cite{wise}, we do not use label smoothing.

\subsection{Validation of CAM-based Object Masking}

In Sec.~4.2, to verify that our CAM-based object masking can effectively mask the patches that cover the main object, we report the average object masking rate and IoU during training with different CAM score thresholds.
Since we do not have the ground truth of the masks of main objects for ImageNet, we approximate it by the prediction of Mask2Former \cite{cheng2022masked}, a segmentation model pre-trained on COCO \cite{coco} (the specific model is reported in \cref{sec:existing_asset}).
We select three super-classes defined in Restricted ImageNet \cite{tsipras2018robustness} that can be recognized by the segmentation model, \ie, Dog, Cat and Bird, which cover 144 ImageNet classes in total.
For each training image of these classes, we obtain the pixel-level segmentation mask $M_{seg}$ corresponding to the super-class, and compare it with our patch-level CAM-based mask, which is translated to a pixel-level mask $M_{\text{CAM}}$ according to the correspondence between patches and pixels. 

The metrics in Tab.~2 in the main text are defined as follows. Formally, a mask $M$ of an image $I$ is defined as a subset of the pixels. Let $n(\cdot)$ denote the number of pixels in a mask or an image. Then, the metrics are defined as:
\begin{itemize}
    \item Image masking rate: $\displaystyle \frac{n(M_{\text{CAM}})}{n(I)}$;
    \item Object masking rate: $\displaystyle \frac{n(M_{\text{CAM}} \cap M_{seg})}{n(M_{seg})}$;
    \item IoU: $\displaystyle \frac{n(M_{\text{CAM}} \cap M_{seg})}{n(M_{\text{CAM}} \cup M_{seg})}$.
\end{itemize}

\subsection{WiSE-KD}

In Sec.~4.4, we consider using the WiSE-FT \cite{wise} model as a teacher model, and add the vanilla knowledge distillation loss \cite{kd} to our training objective, \ie,
\begin{equation}
    \label{eq:objective_with_wise_kd}
    \begin{gathered}
    \mathcal{L} = \mathcal{L}_\text{CE}(g(f(x)),y) + \gamma \mathcal{L}_\text{KL}(g(f(x)), g_e(f_e(x))) \\ + \beta \mathcal{L}_\text{MSE}(\hat{f}(x_{cf}),f(x_{cf})),
    \end{gathered}
\end{equation}
where $\mathcal{L}_\text{KL}$ is the Kullback-Leibler divergence loss, and $f_e$ and $g_e$ are the encoder and classification head of the ensemble model produced by WiSE-FT, correspondingly.
We set $\gamma=1$, and use the WiSE-FT model with $\alpha=0.5$. The temperature of the vanilla knowledge distillation is 10.

\section{Use of Existing Assets}
\label{sec:existing_asset}

\paragraph{Datasets.}

In this paper, we utilize the following existing benchmark datasets without modification or repackaging:
\begin{itemize}
    \item ImageNet \cite{imagenet} (\url{https://www.image-net.org/})
    \item ImageNet-V2 \cite{imagenet_v2} (\url{https://github.com/modestyachts/ImageNetV2})
    \item ImageNet-R \cite{imagenet_r} (\url{https://github.com/hendrycks/imagenet-r})
    \item ImageNet-Sketch \cite{imagenet_s} (\url{https://github.com/HaohanWang/ImageNet-Sketch})
    \item ObjectNet \cite{objectnet} (\url{https://objectnet.dev/})
    \item ImageNet-A \cite{imagenet_a} (\url{https://github.com/hendrycks/natural-adv-examples})
\end{itemize}
In our experiments, we select the hyper-parameters based on validation accuracy on ImageNet, and use the other datasets solely for robustness evaluation. For ObjectNet, we follow the official guidance to remove the red borders of the images before other preprocessing steps in evaluation.

\paragraph{Code and pre-trained model weights.}

The experiments in this paper are based on the code and pre-trained model weights provided by the following packages or GitHub repositories:
\begin{itemize}
    \item PyTorch \cite{paszke2019pytorch} (\url{https://github.com/pytorch/pytorch})
    \item CLIP \cite{clip} (\url{https://github.com/openai/CLIP})
    \item WiSE-FT \cite{wise} (\url{https://github.com/mlfoundations/wise-ft})
    \item Model Soup \cite{model_soup} (\url{https://github.com/mlfoundations/model-soups/issues/1}): we use the pre-trained weights of uniform soup provided by the authors in an issue.
    \item Mask2Former \cite{cheng2022masked} (\url{https://github.com/facebookresearch/Mask2Former/blob/main/MODEL_ZOO.md}): we use the pre-trained model with ID \verb|48558700_7|.
\end{itemize}

\end{document}